\documentclass[10pt, a4paper]{article}
\usepackage{booktabs}

\usepackage[final]{lrec2026} 
\usepackage{soul}       
\usepackage{xcolor}     
\soulregister\cite7
\soulregister\ref7
\soulregister\textit7
\soulregister\textbf7

\title{HealthNLP\_Retrievers at ArchEHR-QA 2026: Cascaded LLM Pipeline for Grounded Clinical Question Answering}

\name{Md Biplob Hosen$^{1,2}$, Md Alomgeer Hussein$^1$, Md Akmol Masud$^2$, \\
      {\bfseries \large Omar Faruque$^1$, Tera L Reynolds$^1$, Lujie Karen Chen$^1$}}

\address{$^1$University of Maryland Baltimore County, Baltimore, MD, USA \\
         $^2$Jahangirnagar University, Savar, Dhaka, Bangladesh \\ 
         {\{mhosen1, mdalomh1, omarf1, reynoter, lujiec\}@umbc.edu}, {masud.stu2018@juniv.edu}
}

\abstract{
Patient portals now give individuals direct access to their electronic health records (EHRs), yet access alone does not ensure patients understand or act on the complex clinical information contained in these records. The ArchEHR-QA 2026 shared task addresses this challenge by focusing on grounded question answering over EHRs, and this paper presents the system developed by the HealthNLP\_Retrievers team for this task. The proposed approach uses a multi-stage cascaded pipeline powered by the Gemini 2.5 Pro large language model to interpret patient-authored questions and retrieve relevant evidence from lengthy clinical notes. Our architecture comprises four integrated modules: (1) a few-shot query reformulation unit which summarizes verbose patient queries; (2) a heuristic-based evidence scorer which ranks clinical sentences to prioritize recall; (3) a grounded response generator which synthesizes professional-caliber answers restricted strictly to identified evidence; and (4) a high-precision many-to-many alignment framework which links generated answers to supporting clinical sentences. This cascaded approach achieved competitive results. Across the individual tracks, the system ranked 1st in question interpretation, 5th in answer generation, 7th in evidence identification, and 9th in answer-evidence alignment. These results show that integrating large language models within a structured multi-stage pipeline improves grounding, precision, and the professional quality of patient-oriented health communication. To support reproducibility, our source code is publicly available in our \href{https://github.com/HosenMB/ArchEHR-QA-2026}{GitHub repository}.
 \\ \newline
}

\begin{document}

\maketitleabstract

\section{Introduction}
Large language models (LLMs) have significantly advanced consumer health question answering. Systems such as Med-PaLM 2 achieve expert-level performance on medical licensing benchmarks \cite{singhal2025toward}. Despite these advances, general-purpose LLMs often struggle to provide grounded answers. Grounding refers to the ability to anchor responses in a specific patient’s medical history without introducing unjustified clinical information \cite{wang2025medical}. This is a major limitation in EHRs, demonstrating a clear vocabulary gap. Patient inquiries are often verbose, informal, and emotionally expressive. The clinical notes remain dense, fragmented, and written in a specialized technical language \cite{nie2014bridging, abacha2019role}. As a result, translating patient-authored questions into clinically meaningful information remains a central challenge for patient-facing health technologies.

The ArchEHR-QA shared task series addresses the communication gap between patient language and clinical documentation in both directions. The task builds on a curated dataset derived from the MIMIC-III critical care database \cite{johnson2016mimic, soni2026dataset}. Unlike earlier clinical question answering tasks focused on extracting facts from large medical corpora \cite{jin2021pubmedqa}, ArchEHR-QA requires systems to operate directly on patient-specific EHR excerpts. Participating systems must interpret subjective patient narratives, identify minimal supporting evidence within complex clinical documentation, and generate tightly constrained grounded answers \cite{soni2026archehr}.

The initial 2025 iteration attracted 29 participating teams submitting 75 systems \cite{soni2025archehr}. This participation reflects strong research interest in evidence-grounded clinical question answering. The 2026 iteration expands the challenge to four complementary subtasks: question interpretation, evidence identification, answer generation, and answer-evidence alignment. Models such as the Gemini series demonstrate strong long-context reasoning capabilities within clinical text \cite{team2024gemini}. Precise alignment between generated answers and sentence-level evidence remains a major challenge \cite{gao2023enabling}. Research increasingly supports multi-stage architectures which decompose complex reasoning tasks into structured intermediate steps instead of relying on single-pass generation \cite{ghafoor2025enhancing}. Role-based prompting and few-shot demonstrations further guide model outputs toward a professional clinical communication style suitable for patient-facing applications \cite{singhal2023large, nori2023capabilities}.

This paper presents the HealthNLP\_Retrievers system, an applied engineering contribution to the ArchEHR-QA 2026 shared task. The system employs a structured multi-stage architecture for grounded clinical question answering over EHRs. Our primary contributions are three integrated design choices: (1) a persona-driven interpretive query reformulation step that runs upstream of retrieval; (2) a Likert-scale, recall-biased evidence scoring heuristic with dynamic fallback tiers to prevent downstream evidence starvation; and (3) a precision-constrained many-to-many answer-evidence alignment framework. Moving beyond standard Retrieval-Augmented Generation (RAG) paradigms \cite{lewis2020rag}, the system deploys Gemini 2.5 Pro \cite{comanici2025gemini} within this four-stage cascaded pipeline to bridge the semantic divide between patient information needs and the objective clinical realities documented in EHRs.

\section{Related Work}

\subsection{Clinical Question Answering}
Automated clinical QA has evolved from structured knowledge-base lookups to open-domain systems capable of reasoning over unstructured text. Early work on consumer health QA focused on summarizing patient questions to bridge the vocabulary gap between nonexpert users and medical professionals \cite{abacha2019role, abacha2019bridging}. PubMedQA \cite{jin2021pubmedqa} introduced a benchmark for biomedical research QA, while more recent efforts have shifted toward patient-specific QA grounded in EHR data \cite{soni2026dataset}. Unlike general health QA, EHR-grounded QA demands that answers are explicitly linked to verifiable clinical evidence, posing unique challenges in retrieval precision and answer fidelity.

\subsection{LLMs for Medical Reasoning}
Large language models have rapidly advanced medical reasoning capabilities. Med-PaLM 2 \cite{singhal2025toward} established expert-level performance on US medical licensing examination style questions, while GPT-4 demonstrated strong diagnostic reasoning across diverse clinical scenarios \cite{nori2023capabilities}. Chain-of-thought (CoT) prompting \cite{wei2022chain} has proven effective for complex clinical reasoning, enabling models to decompose multistep diagnostic logic. Domain-adapted models such as BioBERT \cite{lee2020biobert} have shown that pretraining on biomedical corpora significantly improves performance on clinical NLP tasks. More recently, the Gemini model family \cite{team2024gemini, comanici2025gemini} has demonstrated strong long-context reasoning suited to processing full clinical notes.

\subsection{Evidence Grounding and RAG}
Retrieval-Augmented Generation (RAG) \cite{lewis2020rag} has emerged as a key paradigm for grounding LLM outputs in external knowledge, reducing hallucinations \cite{gao2023enabling}. In clinical settings, iterative RAG frameworks such as i-MedRAG \cite{xiong2024imedrag} allow models to refine queries across multiple retrieval rounds, improving performance on complex medical questions. Standard RAG pipelines retrieve information at the document or passage level, whereas clinical evidence grounding requires sentence-level precision, a gap our Likert-scale scoring approach directly addresses.

\subsection{Prior ArchEHR-QA Systems}
The ArchEHR-QA 2025 shared task \cite{soni2025archehr} revealed several effective design patterns among the 29 participating teams. LAMAR \cite{yoadsanit2025lamar} employed clinically aligned few-shot learning to generate grounded responses from EHRs, while CUNI \cite{lanz2025cuni} investigated whether smaller, more efficient LLMs matched the performance of frontier models on clinical QA. These studies collectively demonstrated that task decomposition and careful prompt engineering often prove more impactful than model scale alone, an insight that informed our cascaded pipeline design.

\section{Methodology}

\begin{figure*}[t]
    \centering
    \includegraphics[width=0.85\textwidth, height=14cm]{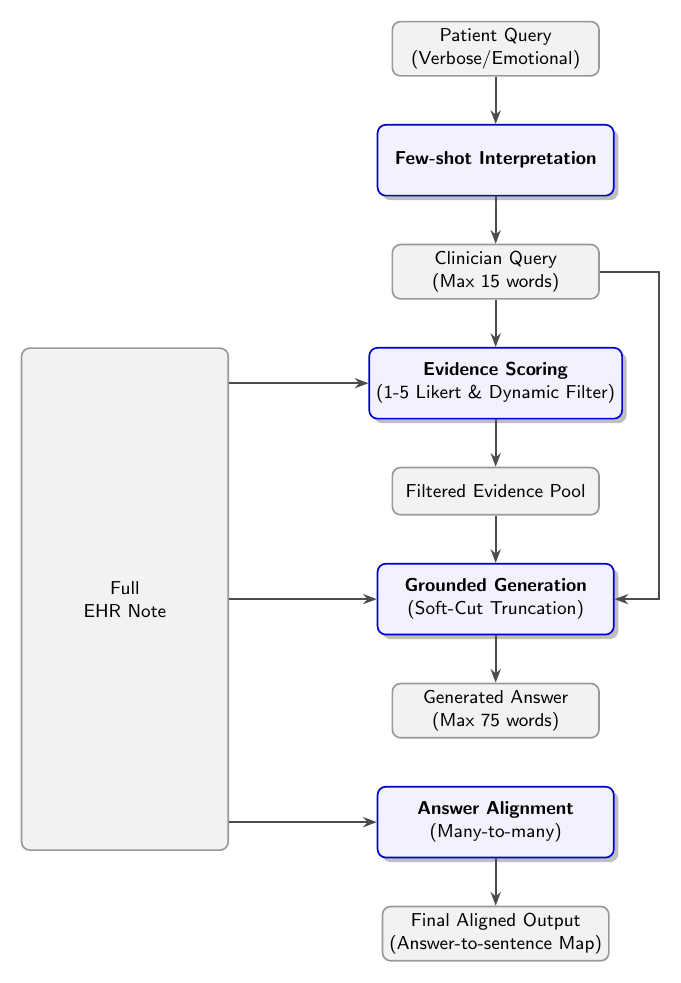}
    \caption{Workflow of the HealthNLP\_Retrievers multi-stage cascaded pipeline.}
    \label{fig:workflow}
\end{figure*}

The proposed HealthNLP\_Retrievers system is a four-stage cascaded pipeline designed to bridge the semantic gap between subjective patient narratives and objective clinical documentation. The system is built using the Gemini 2.5 Pro model to utilize its extended context window to process significantly larger clinical notes. Our approach emphasizes interpretative grounding, interpreting the user's intent ahead of retrieving evidence, and generates a heavily constrained answer. The complete architecture and workflow of this cascaded pipeline are illustrated in Figure~\ref{fig:workflow}.

\subsection{Few-shot Question Interpretation}
Patient-authored questions are frequently characterized by high verbosity and informal phrasing, which introduce significant noise into downstream retrieval tasks, while emotional context often provides important signals about patient concerns and needs \cite{abacha2019role}. To mitigate retrieval noise while preserving clinically meaningful context, we implemented a query interpretation module.

We employ a clinical administrative assistant persona to transform raw narratives into concise, professional queries. The module utilizes a few-shot prompting strategy, constructing a prompt with three examples sampled from the training set. The prompt enforces strict rules including no preambles, standardized usage of pronouns, and a prohibition against diagnostic language, following structured prompting practices used in prior LLM prompting frameworks. To guarantee compliance with the shared task constraints, a post-processing step strictly truncates the generated output to a maximum of 15 words.

\subsection{Heuristic-based Evidence Scoring}
Standard boolean retrieval is insufficient for clinical QA, where evidence is scattered across disjointed note segments. We reformulated evidence identification as a sentence-level scoring task. We instructed the model to act as a clinical evidence scorer and assign a relevance score (1--5) to every sentence, utilizing few-shot examples built from gold-standard annotations. Specifically, the model assigns a score of 5 (critical) to sentences containing direct answers, confirmed diagnoses, or specific medication outcomes. A score of 4 (high context) is allocated to sentences qualifying a critical finding, such as test result context. A score of 3 (relevant) captures background context, admission dates, or clinical history. Finally, scores of 1 and 2 (irrelevant/weak) are reserved for normal findings, administrative headers, or unrelated systems.

\vspace{-1.5ex}
\paragraph{Dynamic Recall-biased Filtering:}
To address the high cost of false negatives in healthcare \cite{burt2017burden}, we implemented a dynamic filtering mechanism. The system first isolates strict evidence (scores $\ge 4$). If no strict evidence is found, it falls back to a lenient set (scores $\ge 3$) to prevent downstream starvation. In the event of an API blockage or formatting error, the system relies on a simple fallback: selecting the first three sentences of the note.

\subsection{Grounded Answer Generation}
The generation module drafts a concise, professional answer capped at 75 words. By prompting the model with a clinical documentation specialist persona, we ensured the final output maintained a strictly objective tone. Specifically, this persona functions as a zero-shot constraint that restricts the model to precise medical terminology, eliminates conversational preambles, and strictly grounds the output to prevent the inclusion of external medical knowledge.

\paragraph{Retrieval-Augmented Generation (RAG):}
The prompt is constructed with three distinct context blocks: (1) the clinical interpretation of the patient question, (2) the explicitly filtered evidence text from stage 2, and (3) the full clinical note as a secondary context. We explicitly instruct the model to prioritize the filtered evidence while the full note serves as a safety net for resolving implicit references---a deliberate "focus-then-expand" design that reduces hallucinations while preventing under-specification. This dual-context design is consistent with the official task input specification \cite{soni2026archehr}.

\paragraph{Soft-cut Truncation Heuristic:}
Given the strict 75-word constraint, we implemented a custom truncation logic. Post-generation, if the text exceeds the limit, it is hard-cut at 75 words. We prefer to truncate the text cleanly at the end of a sentence. The script looks for the last available period, and if cutting there retains at least 60\% of the maximum length, we accept the shorter version. If not, we enforce the hard limit and manually append a period so the output doesn't look broken.

\subsection{Answer-Evidence Alignment}
In the final stage, we tackle many-to-many alignment between the answers and the source text. For the official Subtask 4 evaluation \cite{soni2026archehr}, the system takes pre-segmented clinician answers and the full clinical note as inputs. While a real-world deployment would feed the output of Subtask 3 directly into this stage, we adhered strictly to the shared-task format for testing. We prompted the model to link evidence conservatively, ensuring it only cites sentences that directly back up the answer without pulling in unnecessary background details. To demonstrate this exactness, we embedded a static set of few-shot examples from the development data. The system returns a JSON array that maps every answer sentence to its corresponding evidence sentence indices.

\subsection{Implementation Details}
All experiments were conducted in Python via the Gemini 2.5 Pro API. For tasks requiring strict JSON compliance (Subtasks 1, 2, and 4), the API temperature was set to absolute zero to prevent variation. We slightly increased this to 0.1 for Subtask 3 to help the generated answers read more naturally. Additionally, to ensure the model wouldn't censor valid clinical terminology, we bypassed the standard API safety blocks by setting every \texttt{HarmCategory} to \texttt{BLOCK\_NONE}.

\subsection{Evaluation Metrics}
We evaluated our system using the official metrics defined by the ArchEHR-QA organizers. For the generative tasks (Subtasks 1 and 3), performance was measured using a mix of lexical, semantic, and clinical scores. Specifically, both subtasks relied on ROUGELsum \cite{lin2004rouge}, BERTScore \cite{zhang2019bertscore}, AlignScore \cite{zha2023alignscore}, and MEDCON \cite{yim2023aci} to ensure clinical concepts were preserved accurately. Subtask 3 also included BLEU \cite{papineni2002bleu} and SARI \cite{xu2016optimizing} to evaluate how well the model simplified the text. The official overall score for both generative subtasks is calculated as the unweighted average of their respective constituent metrics.

For the information extraction and classification components (Subtasks 2 and 4), the evaluation relies on standard precision, recall, and F1-scores. Subtask 2 is evaluated comprehensively under both strict and lenient boundary-matching criteria, reporting these metrics at both the micro and macro levels; the strict micro F1-score is designated by the organizers as the overall ranking metric for this subtask. Similarly, Subtask 4 is assessed utilizing micro precision, micro recall, and micro F1-scores. In strict alignment with the official task guidelines, the micro F1-score serves as the overall aggregate metric for Subtask 4.

\section{Result Analysis}
\begin{table*}[t]
\centering
\begin{tabular}{lc|lcc|lcc}
\hline
\textbf{Subtask} & \textbf{Rank} & \textbf{Primary Metric} & \textbf{Ours} & \textbf{Top} & \textbf{Secondary Metric} & \textbf{Ours} & \textbf{Top} \\ \hline
 Question Interpretation & 1 & Overall & 31.2 & 31.2 & BERTScore & 46.8 & 46.8 \\
\hline
Evidence Scoring & 7 & Strict Micro F1 & 60.2 & 63.7 & Lenient Micro F1 & 68.3 & 72.4 \\
\hline
Answer Generation & 5 & SARI & 59.2 & 58.6 & AlignScore & 33.6 & 31.7 \\
\hline
Alignment & 9 & Micro F1 & 76.9 & 81.5 & Micro Precision & 83.8 & 88.0 \\ \hline
\end{tabular}

\vspace{1em}
\caption{Official leaderboard performance of the HealthNLP\_Retrievers system compared to the top-performing system in each subtask.}
\label{tab:results}
\end{table*}

In this section, we analyze the performance of the HealthNLP\_Retrievers pipeline on the ArchEHR-QA 2026 dataset. Our evaluation follows the dual-metric framework of the challenge, focusing on factuality (alignment and evidence identification) and relevance (textual quality) \cite{soni2026archehr}. As shown in Table \ref{tab:results}, our cascaded system demonstrated competitive performance across all four tracks, validating the efficacy of our multi-stage LLM prompting strategy.

\subsection{Interpretative Accuracy}
Our few-shot query interpretation module achieved the 1st place ranking on the ArchEHR-QA leaderboard for Subtask 1, earning an overall score of 31.2. The system demonstrated strong performance across both lexical and semantic evaluations, recording a ROUGELsum score of 35.3 and a BERTScore of 46.8. This secured the top position in a highly competitive field, representing a narrow but decisive improvement over the runner-up team, KPSCMI (overall score of 30.8), and the third-place team, OptiMed (overall score of 29.9).

This outcome highlights the real-world effectiveness of our persona-based prompting strategy. By instructing the model to act as a clinical administrative assistant and enforcing a strict 15-word limit, we successfully stripped away the emotional, verbose parts of the patient's question to leave a tight clinical search query. While other teams likely struggled to cut the word count without dropping critical details, our BERTScore of 46.8—which sits noticeably higher than KPSCMI's 41.0 and OptiMed's 43.1—proves we could aggressively summarize requests without losing their actual meaning. Keeping that core intent intact gave our downstream retrieval module a solid, objective clinical target to search for, rather than getting thrown off by informal patient phrasing.

\subsection{Evidence Identification}
In the evidence extraction phase, our 1--5 Likert-scale scoring approach achieved a strict micro F1 score of 60.2, placing us 7th overall (compared to the top system, Neural, at 63.7). Looking closely at the metrics, our strict micro recall (62.1) noticeably outperformed our strict micro precision (58.4). This result was entirely intentional. We specifically designed the retrieval prompt to lean heavily toward recall to ensure the downstream generator never suffered from evidence starvation. We knew this safety-first approach would cost us some strict precision compared to the top teams, as the model occasionally flagged helpful background context as essential evidence.

The lenient evaluation scores clearly back up this design trade-off. Under the lenient framework, our micro precision jumped 17.3 points to 75.7, pulling our overall lenient micro F1 up to 68.3. This massive gap between the strict and lenient precision tells us exactly what the model was doing: while it pulled in too much text according to the rigid gold-standard boundaries, those extra sentences were still highly relevant and clinically adjacent to the actual answer. By accepting a few more false positives, we successfully kept our false negative rate low. This gave the final text generation stage a much richer set of facts to work with, rather than filling the context window with completely irrelevant noise.

\vspace{-1ex}
\subsection{Grounded Generation}
The generation module achieved 5th place on the ArchEHR-QA leaderboard with an overall score of 34.6. The system's performance was anchored by a strong SARI score of 59.2 and a BERTScore of 43.8. Notably, our SARI score---a metric specifically designed to evaluate text rewriting and simplification---outperformed several higher-ranked teams, including the 1st-place team, WisPerMed (58.6), and the 4th-place team, Neural (57.7). This underscores the effectiveness of our clinical documentation specialist persona and soft-cut truncation heuristic in synthesizing concise, professionally formatted responses that effectively translate complex clinical context into accessible language.

Despite these strong semantic indicators, the relatively low BLEU score of 7.0 highlights a universal vocabulary gap challenge in clinical text generation, mirroring the single-digit BLEU scores observed even among the top-performing teams (e.g., 9.9 for WisPerMed). While our system generated semantically accurate professional responses, the specific lexical phrasing often diverged from the rigid n-gram overlaps expected by the reference gold standard. Crucially, the MEDCON score of 38.7 indicates a robust adherence to medical concept constraints. This proves that our strict RAG-based architecture successfully restricted its generation strictly to the provided evidence pool, synthesizing the necessary clinical information without hallucinating external or unsupported medical concepts.

\begin{table*}[ht]
\centering
\small
\begin{tabular}{lcccc}
\toprule

\multicolumn{5}{c}{\textbf{Subtask 1}} \\ 
\midrule
\textbf{Setting} & \textbf{BERTScore} & \textbf{ROUGELsum} & \textbf{AlignScore} & \textbf{MEDCON (UMLS)} \\ 
\midrule
Zero-shot & 34.13 & 18.75 & 12.80 & 9.20 \\
Few-shot (Ours) & 45.09 & 32.55 & 23.10 & 17.20 \\ 

\midrule
\midrule 

\multicolumn{5}{c}{\textbf{Subtask 2}} \\ 
\midrule
\textbf{Anchor} & \textbf{Strict Micro F1} & \textbf{Strict Micro Precision} & \textbf{Strict Micro Recall} & \textbf{Lenient Micro F1} \\ 
\midrule
Patient Narrative & 60.20 & 50.56 & 74.38 & 63.60 \\
Clinician Query (Ours) & 62.90 & 61.42 & 64.46 & 65.82 \\ 

\bottomrule
\end{tabular}

\vspace{1em}
\caption{Impact of few-shot formulation (Subtask 1) and clinical query anchors (Subtask 2).}
\label{tab:ablation_combined}
\end{table*}

\subsection{Evidence Alignment}
Our alignment module finished in 9th place for Subtask 4 with an overall micro F1 of 76.9. We specifically designed the prompt instructions to be highly selective, which pushed the model to only extract the most certain evidence. This precision-heavy approach resulted in a micro precision score of 83.8. We found this result particularly competitive, as it slightly outperformed or matched top-tier systems like Yale-DM-Lab (83.3) and UIC-AIHealth4All (83.6). These numbers confirm that the module is highly effective at filtering out irrelevant noise and isolating only the most direct supporting sentences from the notes.

On the other hand, our micro recall of 71.1 reflects a specific design choice in our architecture. Because we used such strict prompting heuristics, the model occasionally skipped over sentences that provided implicit support or were part of complex, multi-hop evidence chains. However, we view this high-precision focus as a necessary safeguard for a rigorous verification layer. By favoring exact evidence over a wider contextual window, we ensured the system only cited facts that were explicitly stated in the clinical note. This essentially acts as a final check to prevent the model from making up connections that aren't explicitly stated in the clinical record.

\subsection{Ablation Study}
To evaluate the impact of our primary design choices, we conducted ablation experiments across Subtask 1 and Subtask 2 using the official development dataset, with the results summarized in Table \ref{tab:ablation_combined}. For the question interpretation task (Subtask 1), we tested our few-shot clinical assistant persona against a standard zero-shot baseline. The few-shot setup delivered major improvements across all semantic and clinical metrics. Most notably, the MEDCON score jumped from 9.20 to 17.20. This significant increase in medical entity retention shows that using curated examples is essential for bridging the gap between informal patient language and professional clinical terms.

For evidence identification (Subtask 2), we looked at whether using reformulated clinician queries worked better than using the raw patient narratives as retrieval anchors. While the longer patient narratives naturally led to a higher recall (74.38), they also introduced a lot of noise, which dragged precision down to 50.56. In contrast, our clinician-centered queries achieved a much better balance, raising the Strict Micro F1 to 62.90 and Precision to 61.42. This confirms that a concise, 15-word query is much more effective at filtering out irrelevant patient details while keeping the core clinical facts needed for accurate evidence retrieval.

\subsection{Discussion}
Our results highlight a clear tension between making a query concise and keeping it clinically detailed. In Subtask 1, while we took first place overall, our MEDCON score of 18.7 lagged behind KPSCMI (27.9) and Neural (25.6). This suggests that our administrative assistant persona and 15-word limit were almost too effective at cutting noise—they sometimes stripped away critical medical entities in favor of general descriptions. For example, when a patient described a ``dull to deep pain'' while listing a toxicological history of ``trihexyphenidyl, thorazine, and cocaine,'' the gold-standard query filtered the descriptive fluff but kept the specific drugs. Our system, however, generalized this into: ``What are the potential causes of the patient's persistent chest pain?'' While our version was readable, it lost the specific toxicological context that a human clinician would immediately flag as a high-priority lead. This preference for accessible, general language over technical precision also impacted Subtask 3; while our high SARI scores confirm the model successfully improved text simplicity for a lay audience, this same behavior resulted in a low BLEU score (7.0). This indicates that while our generative outputs remained semantically accurate, the stylistic shift toward simplified vocabulary created a disconnect with the exact n-gram overlaps of the more technically-worded reference labels.

Beyond these vocabulary issues, we saw a clear imbalance between precision and recall in Subtasks 2 and 4. In Subtask 2, our recall-heavy approach pushed precision down to 58.4, while the precision-focused prompting in Subtask 4 limited our recall to 71.1, falling behind teams like OptiMed (79.8). It appears our alignment layer became too exclusionary, often discarding implicit support to verify core facts. In Case 4, for instance, our model correctly identified that the patient had ``low-output, acute-on-chronic heart failure.'' While the system verified the specific procedures and outcomes (Sentences 5, 6, 13, 19, 20), it missed the foundational admission context and diagnostic proof, such as the 25\% ejection fraction and congestive hepatopathy (Sentences 10, 11, 18). This shows that while strict filtering is great for verifying specific treatments, it can severely hurt recall by stripping away the underlying diagnostic reasoning that clinicians rely on.

\vspace{-1ex}
\section{Conclusion}
We presented the HealthNLP\_Retrievers system, a cascaded pipeline for the ArchEHR-QA 2026 shared task. By decomposing clinical QA into query interpretation, heuristic scoring, and precise alignment, our system effectively bridged the gap between patient narratives and clinical notes. This modular design, driven by our few-shot query reformulation strategy, secured 1st place in Question Interpretation (Subtask 1) with an overall score of 31.2 and established a competitive framework for grounded evidence extraction.

Our ablation analysis demonstrated that formulating concise clinical queries significantly improves medical entity retention and retrieval precision. However, error analysis revealed a continuing performance trade-off between concise summarization and medical entity retention, as evidenced by our MEDCON scores. Furthermore, our high-precision alignment strategy in Subtask 4 successfully mitigated hallucinations but resulted in lower recall compared to top-tier competitors. Future work will focus on iterative self-correction to reduce error propagation and the distillation of this pipeline into open-weight models for privacy-preserving clinical deployment.

\section{Limitations}
This work has several limitations. First, the system relies exclusively on the Gemini 2.5 Pro API, introducing a dependency on a proprietary, closed-weight model. This limits reproducibility and adaptability in privacy-sensitive environments, as all clinical text is transmitted to an external server. Second, our evaluation is confined to the ArchEHR-QA 2026 benchmark; performance on other clinical QA datasets or EHR systems remains unexplored. Third, the strictly cascaded architecture propagates errors across stages: an imprecise query reformulation in Subtask 1 can degrade evidence retrieval precision in Subtask 2, which in turn may degrade the quality of the generated answer. Future work should address these gaps through evaluation on open-weight, locally deployable models and the development of end-to-end strategies to mitigate cascaded error propagation.

\section{Ethics Statement}

This work utilizes the ArchEHR-QA dataset \cite{soni2026dataset}, which is derived from the de-identified MIMIC-III database \cite{johnson2016mimic} and accessed through PhysioNet under appropriate data use agreements. To mitigate privacy risks, the patient questions in the dataset are inspired by, rather than directly sourced from, real clinical inquiries. Furthermore, in compliance with both the MIMIC-III data use agreement and ArchEHR-QA 2026 shared task guidelines, we ensured that only strictly de-identified clinical note excerpts were transmitted to the Gemini API. No raw patient identifiers were included in any API calls, and all data processing remained fully compliant with PhysioNet regulations.

We emphasize that the system described herein is designed exclusively as a research prototype and is not intended for direct clinical deployment without rigorous validation by qualified healthcare professionals. For the purposes of this experimental evaluation, all model safety filters were disabled solely to allow the system to accurately process explicit clinical terminology without unwarranted refusal.

\section{Bibliographical References}
\bibliographystyle{lrec2026-natbib}
\bibliography{ref}

\end{document}